\pdfoutput=1

\documentclass[11pt]{article}

\usepackage{ACL2023}

\usepackage{times}
\usepackage{latexsym}

\usepackage[T1]{fontenc}

\usepackage[utf8]{inputenc}

\usepackage{microtype}

\usepackage{inconsolata}

\title{AutoConv: Automatically Generating Information-seeking \\
Conversations with Large Language Models}

\author{
Siheng Li$^{1 \dagger}$\thanks{$^*$ This work is done when Siheng Li is an intern at Huawei Noah's Ark Lab.},
Cheng Yang$^{1}$\thanks{$^\dagger$ Equal contribution.}, 
Yichun Yin$^{2}$, 
Xinyu Zhu$^{1}$,
Zesen Cheng$^{3}$
\\
{\bf Lifeng Shang}$^{2}$, 
{\bf Xin Jiang}$^{2}$, 
{\bf Qun Liu}$^{2}$, 
{\bf Yujiu Yang}$^{1}$\thanks{$^\ddagger$ Corresponding author.} \\ 
$^1$Shenzhen International Graduate School, Tsinghua University \\
$^2$Huawei Noah’s Ark Lab, $^3$Peking University \\
\{lisiheng21, yangc21\}@mails.tsinghua.edu.cn \\
\{yinyichun, shang.lifeng, jiang.xin, qun.liu\}@huawei.com \\
yang.yujiu@sz.tsinghua.edu.cn
}

\usepackage{array}
\usepackage{pifont}
\usepackage{tabularx}
\usepackage{adjustbox}
\usepackage{multirow}
\usepackage{enumitem}
\usepackage{xspace}
\usepackage{tcolorbox}
\usepackage{booktabs,amsfonts,dcolumn}
\usepackage{hyperref}
\usepackage{url}
\usepackage{amsmath,amsthm,amsfonts,amssymb,bm,stmaryrd,bbm}
\usepackage[noorphans,vskip=0.75ex,leftmargin=2ex]{quoting}
\usepackage{graphicx}
\usepackage{enumerate}

\newcommand\ti[1]{\textit{#1}}

\newcommand\tb[1]{\textbf{#1}}

\newcommand\script[1]{{\small \texttt{#1}}}

\newcommand{\vsd}{\vspace{-5pt}}

\definecolor{hong}{HTML}{F6416C}
\definecolor{lan}{HTML}{4D77FF}
\definecolor{lv}{HTML}{00B8A9}
\definecolor{huang}{HTML}{FF8E00}

\newcommand\cmark {\textcolor{green}{\ding{52}}}
\newcommand\xmark {\textcolor{red}{\ding{55}}}

\usepackage{colortbl}
\definecolor{ggray}{RGB}{127,127,127}
\definecolor{aliceblue}{rgb}{0.94, 0.97, 1.0}

\usepackage{arydshln}

\begin{document}
\maketitle

\begin{abstract}
    Information-seeking conversation, which aims to help users gather information through conversation, has achieved great progress in recent years.
    However, the research is still stymied by the scarcity of training data.
    To alleviate this problem, we propose AutoConv for synthetic conversation generation, which takes advantage of the few-shot learning ability and generation capacity of large language models (LLM).
    Specifically, we formulate the conversation generation problem as a language modeling task, then finetune an LLM with a few human conversations to capture the characteristics of the information-seeking process and use it for generating synthetic conversations with high quality. 
    Experimental results on two frequently-used datasets verify that AutoConv has substantial improvements over strong baselines and alleviates the dependence on human annotation.
    In addition, we also provide several analysis studies to promote future research. 
\end{abstract}
\section{Introduction}

In information-seeking conversations, users repeatedly ask questions based on their interests, and the dialogue system provides answers to fulfill their information needs \citep{stede2004information, DBLP:conf/emnlp/ChoiHIYYCLZ18, DBLP:journals/tacl/ReddyCM19}.
This scenario is important for addressing real-world open-ended questions, which requires discussions to explore in depth \citep{DBLP:conf/icml/DaiCZARGG22}, e.g., \ti{How to learn more efficiently}? 
Though great progress has been achieved in recent years, most existing researches depend on abundant human annotation, which can be highly costly and limited in knowledge coverage.

A promising way to alleviate this problem is data augmentation \citep{DBLP:journals/corr/abs-2106-07499}.
Traditional methods, including token-level manipulation \citep{DBLP:conf/naacl/Kobayashi18, DBLP:conf/emnlp/WeiZ19} and sentence-level paraphrasing \citep{DBLP:conf/acl/SennrichHB16}, improve the linguistic diversity of training data.
However, they cannot create conversations grounded on new documents, which are indispensable for dealing with out-of-domain scenarios.
Another line of research focuses on simulation-based methods
\citep{DBLP:journals/corr/abs-2112-08342, DBLP:journals/corr/abs-2205-12609}.
Specifically, they can iteratively generate conversations grounded on new documents based on a span extractor and an utterance generator.
Nevertheless, both the training of the extractor and the generator still require abundant human dialogues.
Besides the above ways,
\citet{DBLP:conf/icml/DaiCZARGG22} propose Dialog Inpainting, which creates information-seeking dialogues by inserting utterances between neighboring sentences in documents. 
One potential risk is the gap between the structure of documents and that of conversations.
Documents are tighter, while real-world conversations are more open-ended.

\begin{table}[t]
\centering
\setlength{\tabcolsep}{1mm}        
\renewcommand{\arraystretch}{1.15}  
\footnotesize
\begin{tabular}{lccc}
\noalign{\hrule height 1.5pt}
   \tb{Method} & \tb{DG} & \tb{Data Needs} \\
\hline
    EDA \citep{DBLP:conf/emnlp/WeiZ19} & \xmark & - \\
    Back-Translation \citep{DBLP:conf/acl/SennrichHB16} & \xmark & - \\
    SeemSeek \citep{DBLP:journals/corr/abs-2205-12609} & \cmark & Large \\
    Dialog Inpainting \citep{DBLP:conf/icml/DaiCZARGG22} & \cmark & Large \\
    \hdashline
    \rowcolor{aliceblue!60}
    \textbf{AutoConv}~(Ours) & \cmark & \tb{Few} \\
\noalign{\hrule height 1.5pt}
\end{tabular}
\caption{
    The differences between AutoConv and others. DG represents whether the augmentation is document grounded, and Data Needs denotes the scale of human conversations used for augmentation. 
}
\label{tab:compare_with_other_methods}
\vspace{-10pt}
\end{table}

To alleviate the above issues, we propose a simple yet effective method \tb{AutoConv} for \tb{Auto}matically generating information-seeking \tb{Conv}ersations, which takes advantage of the few-shot learning ability and generation capacity of large language models (LLM) \citep{DBLP:conf/nips/BrownMRSKDNSSAA20}.
Specifically, we formulate conversation generation as a language modeling task and utilize an LLM for generating synthetic conversations grounded on external documents.
Surprisingly, finetuning with a few human dialogues can help LLM capture the characteristics of the information-seeking process (e.g., grounding, question answering) and generate high-quality synthetic conversations.
Then, we can train a small task model with these dialogues.
The differences between AutoConv and others are shown in Table~\ref{tab:compare_with_other_methods}.

We conduct comprehensive experiments on two frequently-used datasets QuAC \citep{DBLP:conf/emnlp/ChoiHIYYCLZ18} and CoQA \citep{DBLP:journals/tacl/ReddyCM19} in the low-resource setting, where only dozens of human dialogues are available.
The results show that AutoConv has substantial improvements over several strong baselines.
When scaling up the synthetic dialogues, AutoConv has the improvement of up to 5.06 F1 gain compared with directly finetuning, and thus largely reduces the labor force for annotation.
In addition, we find that the small task model trained with synthetic dialogues can even surpass finetuned LLM with only $1.7\%$ parameters.
Moreover, we also investigate the impact of decoding strategy and scaling laws for AutoConv.

\section{Method}
\begin{figure}
    \centering
    \includegraphics[width=1.0\linewidth]{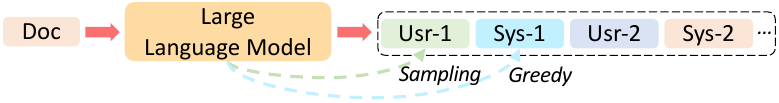}
    \caption{The generation process of AutoConv. We use nucleus sampling for generating user questions and greedy search for generating system answers.
    }
    \label{fig:method}
    \vspace{-5pt}
\end{figure}

\subsection{Task Formulation}
Our goal is automatically generating information-seeking conversations.
Specifically, each conversation is grounded on a document \tb{\ti{d}} and consists of a series of user questions and system answers. 

\subsection{Conversation Generation}
\label{sec:dialogue_generation}

\paragraph{Training.}
We formulate conversation generation as a language modeling task and finetune\footnote{In our preliminary experiments, we try to prompt LLM without training. However, we find that the performance is poor and LLM fails to generate conversations grounded on the documents, similar to the observation in \citet{DBLP:journals/corr/abs-2202-13047}.} an LLM with a few human dialogues (e.g., 50 from QuAC \citep{DBLP:conf/emnlp/ChoiHIYYCLZ18}) to capture the characteristics of information-seeking conversations (e.g., grounding, question answering).
The objective is the negative log-likelihood of each utterance:
\begin{equation}
   \mathcal{L} = - \sum_{t=1}^T\sum_{l=1}^L \log P(\ti{u}_{l}^t|\ti{u}_{<l}^t, \tb{\ti{h}}_{<t}, \tb{\ti{d}}), \nonumber
\end{equation}
where $\tb{\ti{u}}$ represents a user question or a system answer, $\tb{\ti{h}}$ is the dialogue history, $L$ and $T$ are the number of tokens and turns respectively.

\paragraph{Generating.}
Based on the finetuned LLM, we can generate synthetic dialogues with unlabeled documents, as in Figure \ref{fig:method}.
In information-seeking scenarios, user questions are typically open-ended. Thus we choose nucleus sampling \citep{DBLP:conf/iclr/HoltzmanBDFC20} for generating user questions, which has shown great performance in various open-ended generation tasks \citep{DBLP:journals/corr/abs-2202-06417}.
However, when applying a sampling decoding strategy for system answer generation, we find it results in the ``hallucination'' problem \citep{DBLP:conf/emnlp/0001PCKW21}, where the generation is plausible but factually incorrect based on the document.
To this end, we utilize greedy search for answer generation.
Neural language models often generate the same sentences repetitively \citep{DBLP:journals/corr/abs-2206-02369}. 
To alleviate this problem, we first compute the diversity score of each synthetic dialogue as in \citet{DBLP:journals/corr/abs-2202-06417}, which considers the repetition at different $n$-gram levels.
Then, we filter out dialogues based on this score.

After that, a two-stage training strategy is adopted  \cite{DBLP:conf/cvpr/XieLHL20} for training a small task model.
Specifically, we first pre-train it on the synthetic dialogues, then  finetune it on the human dialogues used for finetuning the LLM.
More training details are given in Appendix \ref{app:implementation_details}.
\section{Experiments}
We conduct experiments on QuAC \citep{DBLP:conf/emnlp/ChoiHIYYCLZ18} and CoQA \citep{DBLP:journals/tacl/ReddyCM19}, more details about them are shown in Appendix \ref{app:datasets}.

\subsection{Implementation}
We focus on the low-resource setting, where human dialogues are scarce.
To simulate this setting, we randomly sample a few human dialogues from the training set of QuAC or CoQA, and use them for finetuning the LLM.
We use OPT-13B \citep{DBLP:journals/corr/abs-2205-01068} as the LLM and UnifiedQA-V2-base (222M) \citep{DBLP:journals/corr/abs-2202-12359} as the small task model.
All data augmentation methods use the same training strategy and small task model.
More implementation details are shown in Appendix \ref{app:implementation_details}.


\begin{table*}[t]
\begin{center}
\setlength{\tabcolsep}{4mm}        
\renewcommand{\arraystretch}{1.3}  
\centering
\small
\resizebox{1.0\textwidth}{!}{%
\begin{tabular}{llllcll}
\noalign{\hrule height 1.5pt}
\multirow{2}{*}{\tb{Method}} & &  \multicolumn{2}{c}{\tb{QuAC}} &  & \multicolumn{2}{c}{\tb{CoQA}} \\
\cline{3-7}
& & \multicolumn{1}{c}{\tb{F1}} & \multicolumn{1}{c}{\tb{EM}} & & \multicolumn{1}{c}{\tb{F1}} & \multicolumn{1}{c}{\tb{EM}} \\
\hline
\hline
\rowcolor{ggray!20}
\multicolumn{7}{c}{\it{Prompting}}\\
\hline
GPT-3 Zero-shot \citep{DBLP:conf/nips/BrownMRSKDNSSAA20} &  & 41.5 & - & & 81.5 & - \\
GPT-3 Few-shot \citep{DBLP:conf/nips/BrownMRSKDNSSAA20} & & 44.3 & - & & 85.0 & - \\
\hline
\rowcolor{ggray!20}
\multicolumn{7}{c}{\it{Data Augmentation (50 Human Dialogues)}}\\
\hline
    Finetuning & & 46.57\scriptsize{$\pm$1.29} & 30.68\scriptsize{$\pm$1.25}  & & 70.41\scriptsize{$\pm$0.46} & 60.43\scriptsize{$\pm$0.56} \\
    Back-Translation \cite{DBLP:conf/acl/SennrichHB16} & &  47.92\scriptsize{$\pm$0.49} & 28.26\scriptsize{$\pm$1.39} & & 67.59\scriptsize{$\pm$2.73} & 56.34\scriptsize{$\pm$3.41} \\
    EDA \citep{DBLP:conf/emnlp/WeiZ19} & & 46.04\scriptsize{$\pm$1.28} & 28.88\scriptsize{$\pm$2.20} & & 58.89\scriptsize{$\pm$2.08} & 47.64\scriptsize{$\pm$2.14} \\
    Utterance Manipulation \citep{DBLP:conf/emnlp/ChenY21a} & &  48.83\scriptsize{$\pm$0.63} & 33.91\scriptsize{$\pm$0.73} & & 68.69\scriptsize{$\pm$0.85} & 58.30\scriptsize{$\pm$1.21} \\
    Dialog Inpainting \citep{DBLP:conf/icml/DaiCZARGG22} & & 48.33\scriptsize{$\pm$1.24} & 32.23\scriptsize{$\pm$1.55} & & 70.25\scriptsize{$\pm$0.93} & 59.83\scriptsize{$\pm$0.98} \\
    \hdashline
    \rowcolor{aliceblue!60}
    \textbf{AutoConv} & & \tb{50.48\scriptsize{$\pm$0.94}} & \tb{34.12\scriptsize{$\pm$0.93}} & & \tb{73.87\scriptsize{$\pm$0.85}} & \tb{63.78\scriptsize{$\pm$1.01}} \\
\hline
Human Annotation & & 53.24\scriptsize{$\pm$0.28} & 36.85\scriptsize{$\pm$0.35} & & 76.02\scriptsize{$\pm$0.71} & 65.92\scriptsize{$\pm$1.01} \\
\hline
\rowcolor{ggray!20}
\multicolumn{7}{c}{\it{Data Augmentation (100 Human Dialogues)}}\\
\hline
    Finetuning & & 48.98\scriptsize{$\pm$1.16} & 31.98\scriptsize{$\pm$1.09}  & & 72.78\scriptsize{$\pm$0.69} & 62.41\scriptsize{$\pm$0.85} \\
    Back-Translation \cite{DBLP:conf/acl/SennrichHB16} & &  48.41\scriptsize{$\pm$0.96} & 28.10\scriptsize{$\pm$2.51} & & 69.18\scriptsize{$\pm$2.82} & 57.72\scriptsize{$\pm$3.28} \\
    EDA \citep{DBLP:conf/emnlp/WeiZ19} & & 46.86\scriptsize{$\pm$0.61} & 29.14\scriptsize{$\pm$1.71} & & 60.61\scriptsize{$\pm$4.23} & 49.24\scriptsize{$\pm$4.74} \\
    Utterance Manipulation \citep{DBLP:conf/emnlp/ChenY21a} & &  49.07\scriptsize{$\pm$1.06} & 31.77\scriptsize{$\pm$1.86} & & 69.23\scriptsize{$\pm$0.21} & 59.15\scriptsize{$\pm$0.74} \\
    Dialog Inpainting \cite{DBLP:conf/icml/DaiCZARGG22} & & 49.48\scriptsize{$\pm$0.34} & 33.29\scriptsize{$\pm$0.98} & & 72.15\scriptsize{$\pm$0.74} & 61.80\scriptsize{$\pm$0.99} \\
    \hdashline
    \rowcolor{aliceblue!60}
    \textbf{AutoConv} & & \tb{51.21\scriptsize{$\pm$1.02}} & \tb{34.65\scriptsize{$\pm$1.00}} & & \tb{74.84\scriptsize{$\pm$0.24}} & \tb{64.36\scriptsize{$\pm$0.46}} \\
\hline
Human Annotation & & 54.22\scriptsize{$\pm$0.90} & 37.42\scriptsize{$\pm$2.06} & & 76.35\scriptsize{$\pm$0.51} & 65.71\scriptsize{$\pm$0.55} \\
\noalign{\hrule height 1.5pt}
\end{tabular}
}
\end{center}

\caption{
    Comparison with baselines. All experiments are performed 4 runs with different random seeds. Finetuning means directly training with only human dialogues. All data augmentation methods use the same human dialogues and the same number of synthetic dialogues for the sake of fairness (5 times the number of human dialogues). Human annotation represents replacing the synthetic dialogues with the same number of human dialogues.
}
\label{tab:main_results}
\vspace{-5pt}
\end{table*}

\subsection{Comparison with Baselines}
\label{sec:comparison_with_baselines}

We compare AutoConv with a series of baselines, and the details of them are given in Appendix \ref{app:baselines}.
As shown in Table \ref{tab:main_results}, 
AutoConv achieves better performance than GPT-3 prompting on QuAC with only $0.13\%$ parameters and 50 human dialogues, but is less competitive on CoQA.
We conjecture the reason stems from the intrinsic difference between the two datasets.
CoQA contains more factoid questions, and the answers are named entities or short noun phrases like those in SQuAD \citep{DBLP:conf/emnlp/RajpurkarZLL16}.
By training on large-scale text corpus from a web forum, GPT-3 might implicitly learn the format and structure of question answering \citep{DBLP:conf/iclr/SanhWRBSACSRDBX22}, and thus gets excellent performance on CoQA.
On the other side, QuAC has more open-ended and exploratory questions as in natural conversations, and $86\%$ questions are contextual \citep{DBLP:conf/emnlp/ChoiHIYYCLZ18}.
Therefore, it brings more difficulties for GPT-3 inference with few demonstrations, while our method learns better from both human dialogues and synthetic dialogues.

\begin{figure}
    \centering
    \includegraphics[width=1.0\linewidth]{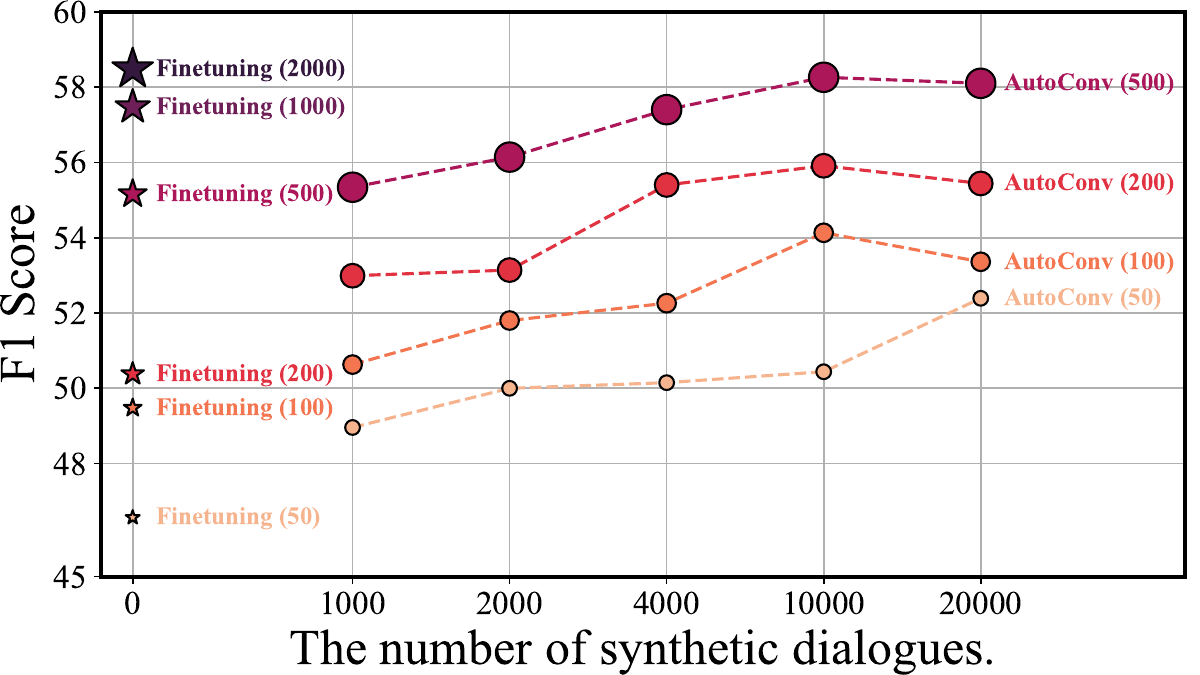}
    \caption{The results of scaling up human dialogues and synthetic dialogues on QuAC. 
    The number in the parentheses represents the number of human dialogues.
    }
    \label{fig:scaling_data}
    \vspace{-5pt}
\end{figure}

Compared with data augmentation methods, AutoConv achieves the best performance on both datasets and mitigates the gap between synthetic dialogues and human upper bounds.
We find that the token-level augmentation method EDA and the sentence-level augmentation method Back-Translation even hurt the performance, which is similar to the observation in \citet{DBLP:journals/corr/abs-2106-07499}. 
One possible reason is that they bring too much noise.
Dialog Inpainting \citep{DBLP:conf/icml/DaiCZARGG22} gets ordinary performance, and the reason possibly derives from the gap between the structure of natural conversations and that of the documents used for constructing synthetic dialogues.

\subsection{Scaling up Human Dialogues and Synthetic Dialogues}
\label{sec:scaling_up_human_dialogues_and_synthetic_dialogues}
In this part, we further analyze the performance of AutoConv when scaling up the human dialogues and synthetic dialogues.
As shown in Figure \ref{fig:scaling_data}, the performance boosts when more human dialogues or synthetic dialogues are used.
With 50 human dialogues, AutoConv outperforms the results of finetuning with 200 human dialogues.
With 500 human dialogues, AutoConv gets competitive performance compared with finetuning with 2000 human dialogues.
These results verify the high quality of synthetic dialogues, and our AutoConv can largely alleviate the labor force for annotation.

\subsection{Comparison with Finetuned Large Language Model}
\begin{table}[t]
\begin{center}
\centering
\setlength{\tabcolsep}{2mm}        
\renewcommand{\arraystretch}{1.4}  
\small
\resizebox{\linewidth}{!}{%
\begin{tabular}{l|cc|cc}
\noalign{\hrule height 1.5pt}
\tb{Model} & \tb{\#Params} & \tb{\#FLOPs} & \tb{F1 (50)} & \tb{F1 (200)}\\
\hline
    Finetuning (LLM) & 12.9B & 7049.3B & 
    \tb{53.53} & 54.85\\
    Finetuning (STM) & \tb{222M} & \tb{60.2B} & 47.97 & 50.38 \\
    \hdashline
    \rowcolor{aliceblue!60}
    \textbf{AutoConv} (STM) & \tb{222M} & \tb{60.2B} & 52.40 & \tb{55.44}\\
\noalign{\hrule height 1.5pt}
\end{tabular}
}
\end{center}
\caption{
    Comparison results on QuAC. 
    Finetuning means training with only human dialogues. 
    AutoConv uses the same human dialogues and 20K synthetic dialogues. 
    LLM is large language model and STM is small task model. 
    The number in the parentheses represents the number of human dialogues.
}
\label{tab:compare_with_finetuning}
\vspace{-5pt}
\end{table}
AutoConv is a kind of symbolic knowledge distillation \citep{DBLP:conf/naacl/WestBHHJBLWC22}, where the finetuned large language model (LLM) transfers its knowledge to the small task model (STM) by generating synthetic dialogues for the training of STM.
Here, we further investigate the effectiveness of AutoConv from the aspect of knowledge distillation.
As shown in Table \ref{tab:compare_with_finetuning}, finetuned LLM has substantial improvements over finetuned STM.
However, it brings large memory and computation cost.
On the other side, our AutoConv not only keeps the efficiency of STM, but also boosts the performance.
Surprisingly, AutoConv even outperforms its teacher model in the 200 human dialogues setting.
Similar observations are found in \citet{DBLP:conf/naacl/WestBHHJBLWC22, DBLP:journals/corr/abs-2202-07922}, while they focus on different tasks.
We leave the analysis of this novel observation for future work.


\subsection{Impact of Decoding Strategy}
\label{sec:decoding_strategy}

\begin{table}[t]
\begin{center}
\centering
\setlength{\tabcolsep}{2mm}        
\renewcommand{\arraystretch}{1.25}  
\small
\resizebox{1.0\linewidth}{!}{%
\begin{tabular}{l|cc}
\noalign{\hrule height 1.5pt}
   \tb{Decoding Strategy} & \tb{F1} & \tb{Exact Match} \\
\hline
Nucleus Sampling ($p=0.8$) & 50.77 & 32.63 \\
Nucleus Sampling ($p=0.9$) & 49.88 & 31.57 \\
Greedy Search & 53.53 & 36.38 \\
Beam Search ($b=4$) & \tb{54.43} & 38.64 \\
\textbf{Beam Search} ($b=8$) & \tb{54.43} & \tb{38.70} \\
\noalign{\hrule height 1.5pt}
\end{tabular}
}
\end{center}
\caption{
    The results of LLM with different decoding strategies for answer generation on QuAC, 50 human dialogues are used for finetuning the LLM. 
}
\label{tab:decoding_strategy}
\vspace{-5pt}
\end{table}

During our preliminary experiments, we find that the decoding strategy is important for system answer generation.
More precisely, we evaluate the answer generation performance of LLM with different decoding strategies on QuAC, and the results are shown in Table \ref{tab:decoding_strategy}.
Though nucleus sampling \citep{DBLP:conf/iclr/HoltzmanBDFC20} has shown great performance in various generation tasks \citep{DBLP:journals/corr/abs-2202-06417}, it performs less competitively than maximization-based decoding strategies for answer generation. 
Compared with beam search, greedy search shows competitive performance and is more efficient. Thus we use greedy search by default in this paper.

\subsection{Scaling Laws}
\label{sec:scaling_laws}
\begin{figure}
    \centering
    \includegraphics[width=\linewidth]{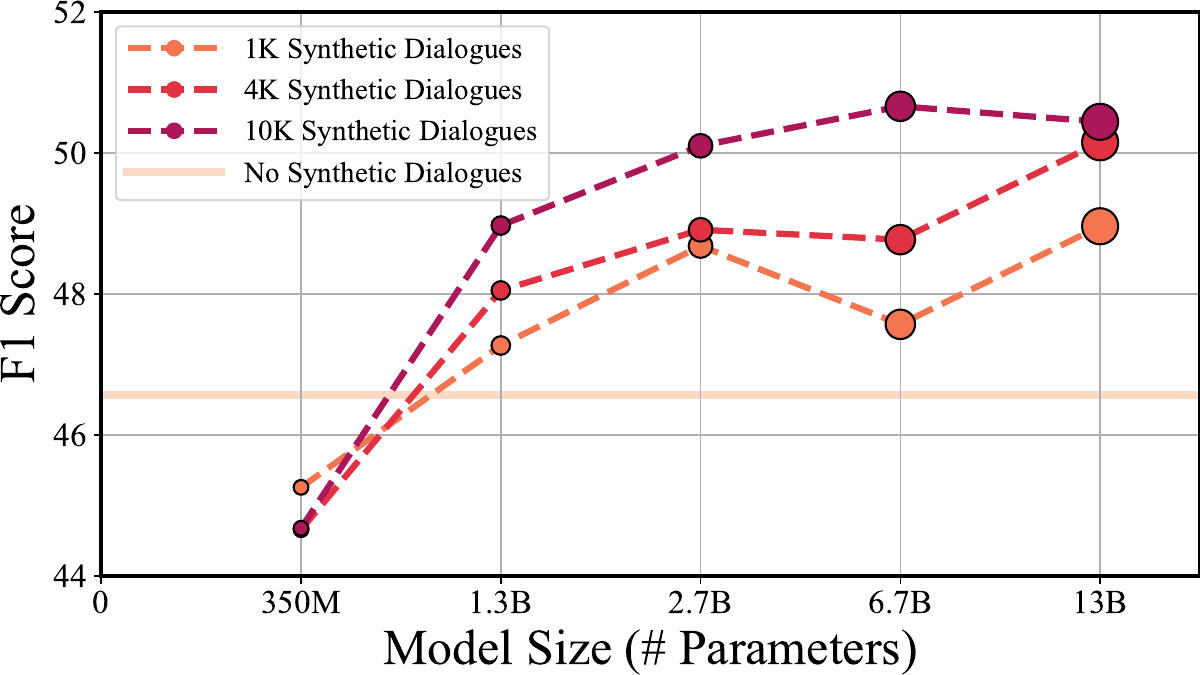}
    \caption{The results of AutoConv with different LLM on QuAC.
    We use different scale of OPT \citep{DBLP:journals/corr/abs-2205-01068} as the LLM.
    All models are trained with 50 human dialogues for fairness and  synthetic dialogues are generated with the corresponding LLM.
    }
    \label{fig:scaling_model}
    \vsd
    \vsd
\end{figure}
We further analyze how the benefit of AutoConv is affected by the scale of LLM. 
As shown in Figure \ref{fig:scaling_model}, the performance gets better with a larger model across a various number of synthetic dialogues.
In addition, when the LM is small (350M) and with limited generation ability, the synthetic dialogues can even hurt the performance when the available human dialogues are scarce.
Due to the limitation of computational resources, we limit our investigation to 13B parameters and leave larger models for future work.

\subsection{Case Study}
\label{sec:case_study}
\begin{table}[htbp]
\centering
\setlength{\tabcolsep}{0mm}        
\renewcommand{\arraystretch}{1.15}  
\begin{tabular}{m{7.7cm}}
\noalign{\hrule height 1.5pt}
\multicolumn{1}{c}{\textbf{Title}} \\
\hline
2006-2007: Ciara: The Evolution and acting debut \\ 

\hline
\multicolumn{1}{c}{\textbf{Document}} \\ 
\hline
  On December 5, 2006, Ciara released her second studio album, Ciara: The Evolution. According to the singer, the title of the album is \textbackslash{}"about so much more than just my personal growth - it's about the evolution of music, the evolution of dance, the evolution of fashion \textbackslash{}" The source of the album\'s creativity such as the sound and edge comes from Ciara in general. Ciara: The Evolution became Ciara\'s first and only number one album on the U.S. Billboard 200, and her second number one on the Top R\&B/Hip-Hop Albums charts with sales of 338,000 in the first week \dots \\

\hline
\multicolumn{1}{c}{\textbf{Dialogue}} \\ 
\hline
\textbf{Usr}:\;What was the evolution?                           \\ 
\textbf{Sys}:\;On December 5, 2006, Ciara released her second studio album, Ciara: The Evolution.\\ 
\textbf{Usr}:\;What was the name of the album?                                                   \\ 
\textbf{Sys}:\;The title of the album is \textbackslash{}"about so much more than just my personal growth - it's about the evolution of music, the evolution of dance, the evolution of fashion.\textbackslash{}" \\
\textbf{Usr}:\;Did the album do well? \\
\textbf{Sys}:\;The album became Ciara's first number one album on the U.S. Billboard 200 and second on the Top R\&B/Hip-Hop Albums charts with sales of more than 338,000.\\
\textbf{Usr}:\;What awards did she win if any?\\
\textbf{Sys}:\;The album went on to be certified platinum by the RIAA in the United States, within only five weeks of its release, and has sold 1.3 million copies according to Nielsen SoundScan.\\
\textbf{Usr}:~What else did she accomplish during that time?\\
\textbf{Sys}:~In support of the album, Ciara went on her first headlining tour in October 2006.\\
\textbf{Usr}:~How long was the tour?\\
\textbf{Sys}:~The tour went to seventeen different clubs in cities throughout the United States.\\
\textbf{Usr}:~Did she have success with any songs?\\
\textbf{Sys}:~The lead single, \textbackslash{}"Get Up\textbackslash{}", which features Chamillionaire, reached number seven in the United States and gained a platinum accreditation.\\
\noalign{\hrule height 1.5pt}
\end{tabular}
\caption{An example of the synthetic conversation generated by AutoConv, the LLM is finetuned with 50 human dialogues from QuAC \cite{DBLP:conf/emnlp/ChoiHIYYCLZ18}.}
\label{tab:case_study_single_column}
\end{table}
In Table~\ref{tab:case_study_single_column}, we present an example of our synthetic conversation for the case study.
The original document describes the singer Ciara's second studio album and her acting debut.
The conversation consists of seven user questions and seven system answers, covering the title and sales of the album, the duration of the tour, etc.
As we can see from this example, the user questions are diverse (e.g. what, how, did, etc.) and the conversation is informative and conversational. 
For example, when the system mentions ``tour'' (the fifth system utterance), the user follows by asking ``How long was the tour?''.

\subsection{Error Analysis}
\label{sec:error_analysis}
To further analyze the limitation of our method, we conduct an error analysis by manually investigating 50 synthetic conversations generated by AutoConv, which is finetuned with 50 human conversations from QuAC \citep{DBLP:conf/emnlp/ChoiHIYYCLZ18}.
Particularly, we find that only 5\% generated questions are not suitable (e.g., misspelled names).
The reason stems from the open-ended characteristic of natural conversation that many kinds of user questions are possible under the same context.  
However, nearly 40\% of system answers are not perfect, and we summarize the wrong answers into four major classes:
\textbf{(1) Irrelevant}: 75\% of them are totally irrelevant to user questions.
\textbf{(2) Related but not Accurate}: 14\% of them contain related knowledge from the grounded documents, but the answers are not accurate.
Take an example in Table~\ref{tab:case_study_single_column}, the second user question asks for the name of the album, which is \ti{Ciara: The Evolution} according to the document.
While the LLM generates the interpretation of the album name by mistake. 
\textbf{(3) Missing}: 4\% of them belong to the missing error that the system answers are ``No Answer'', while the questions actually can be answered based on the documents.
\textbf{(4) Hallucination}: 3\% of them mention hallucination knowledge, which cannot be found in the documents. 
In addition, we also notice that AutoConv is more likely to generate wrong answers when grounding on longer and more complex documents.

\section{Conclusion}
In this paper, we propose a simple yet effective method, AutoConv, which formulates the conversation generation problem as a language modeling task.
Then, based on a large language model and a few human dialogues, AutoConv can generate synthetic dialogues with high quality.
Experimental results on both QuAC and CoQA verify the effectiveness of AutoConv, which alleviates the human efforts for annotation largely.
Furthermore, we also provide case study and error analysis to prompt future research.

\section*{Limitations}
In this paper, we propose a method named AutoConv, which means automatically generating information-seeking conversations with large language models (LLM). 
Though it has achieved great performance on both QuAC \citep{DBLP:conf/emnlp/ChoiHIYYCLZ18} and CoQA \citep{DBLP:journals/tacl/ReddyCM19}, there are still some limitations that should be noticed.

\paragraph{Limitation of LLM.} In our experiments, we use OPT-13B \citep{DBLP:journals/corr/abs-2205-01068} as the LLM for generating synthetic conversations due to the limited computational resources.
Larger models should be considered to further understand the potential ability of AutoConv, e.g., GPT-3 \citep{DBLP:conf/nips/BrownMRSKDNSSAA20}, OPT-175B \citep{DBLP:journals/corr/abs-2205-01068}, BLOOM-176B \citep{DBLP:journals/corr/abs-2211-05100}, and GLM-130B \citep{DBLP:journals/corr/abs-2210-02414} etc.

\paragraph{Limitation of Implementation.} As mentioned in Section \ref{sec:dialogue_generation} and Appendix \ref{app:implementation_details}, our method needs to finetune LLM and generate massive synthetic conversations based on the finetuned LLM, which has a high cost for implementation.

\paragraph{Limitation of Synthetic Dialogues.} As shown in Table \ref{tab:main_results} and Section \ref{sec:error_analysis}, there is still a gap between our synthetic dialogues and human dialogues.
It is important to improve the quality of synthetic dialogues so that we can further alleviate the dependence on human annotation.

\section*{Ethics Statement}
AutoConv is based on large language models (LLM), while LLM has some potential risks, e.g., social bias \citep{DBLP:conf/icml/LiangWMS21}, offensive content \citep{DBLP:journals/corr/abs-2209-07858} etc.
Fortunately, we finetune the LLM to capture the characteristics of the information-seeking process, and the generated conversations are mostly grounded on the provided documents (take an example in Table \ref{tab:case_study_single_column}).
Therefore, our method alleviates the potential risks of directly using LLM.
According to our manual check in error analysis (Section \ref{sec:error_analysis}), we do not find any harmful content in the synthetic conversations.
In addition, we also encourage considering more safety methods \citep{DBLP:journals/corr/abs-2010-07079, DBLP:conf/acl/0012XDCZZP0H22} to guarantee the quality of synthetic conversations.

\section*{Acknowledgements}
This work was partly supported by the National Key Research and Development Program of China (No. 2020YFB1708200) ,  the "Graph Neural Network Project" of Ping An Technology (Shenzhen) Co., Ltd. and AMiner.Shenzhen SciBrain fund.

\bibliography{anthology,custom}
\bibliographystyle{acl_natbib}

\appendix
\clearpage

\section{Datasets}
\label{app:datasets}
\paragraph{QuAC.} QuAC \citep{DBLP:conf/emnlp/ChoiHIYYCLZ18} is a leading conversational question answering dataset, consists of 14K information-seeking dialogues.
Different from the factoid questions in most existing QA datasets, the questions in QuAC are more open-ended and exploratory.
In addition, $86\%$ of questions are contextual, and the model needs to understand the dialogue context to resolve coreference.
As the test set is only available in the QuAC challenge\footnote{\url{https://quac.ai/}}, we evaluate the performance on the development set.

\paragraph{CoQA.} CoQA \citep{DBLP:journals/tacl/ReddyCM19} consists of 127K conversational QA pairs across seven domains. 
Different from QuAC, CoQA focus more on factoid questions, and the answers are mostly named entities or short phrases as in SQuAD \citep{DBLP:conf/emnlp/RajpurkarZLL16}. 
The test set of CoQA is only available in the CoQA challenge\footnote{\url{https://stanfordnlp.github.io/coqa/}}, therefore we evaluate the performance on the development set.

\section{Implementation Details}
\label{app:implementation_details}
\paragraph{General Setting.} All experiments are based on Transformers\footnote{\url{https://huggingface.co/docs/transformers/index}} \cite{DBLP:conf/emnlp/WolfDSCDMCRLFDS20}, DeepSpeed\footnote{\url{https://github.com/microsoft/DeepSpeed}} \cite{DBLP:conf/kdd/RasleyRRH20} and Pytorch Lightning\footnote{\url{https://github.com/Lightning-AI/lightning}}.
We use UnifiedQA-V2-base\footnote{\url{https://huggingface.co/allenai/unifiedqa-v2-t5-base-1363200}} \citep{DBLP:conf/emnlp/KhashabiMKSTCH20, DBLP:journals/corr/abs-2202-12359} as the small task model, which is based on T5 architecture with 222M parameters and pre-trained on many QA tasks (the tasks in our experiments are not included in).
The training of the small task model follows the original paper \citep{DBLP:conf/emnlp/KhashabiMKSTCH20} in a Text-to-Text framework \citep{DBLP:journals/jmlr/RaffelSRLNMZLL20}.
The input is \script{Dialogue History \textbackslash n Document} and the output is \script{System Answer}.

For the training hyperparameters, we set the learning rate as $3e-4$, batch size as $32$, and use Adam optimizer \citep{DBLP:journals/corr/KingmaB14} with warmup learning rate schedule, the warmup ratio is $0.1$.
When comparing with baseline methods as in Section \ref{sec:comparison_with_baselines}, all methods use the same small task model, the same two-stage training strategy \cite{DBLP:conf/cvpr/XieLHL20, DBLP:conf/emnlp/ChenY21a}, the same human dialogues and the same number of synthetic dialogues for fairness (5 times the number of human dialogues).
For the 50 human dialogues setting, we train each model for 1K gradient steps in the pre-training stage and 200 gradient steps in the fintuning stage. 
For the 100 human dialogues setting, the steps are 2K and 400 respectively. 
When scaling up the number of synthetic dialogues as in Section \ref{sec:scaling_up_human_dialogues_and_synthetic_dialogues} and Section \ref{sec:scaling_laws}, the numbers of pre-training steps scale up, which are 2K, 4K, 8K, 20K and 40K for 1K, 2K, 4K, 10K and 20K synthetic dialogues respectively, and the finetuning steps are 200, 400, 800 and 2K for 50, 100, 200 and 500 human dialogues respectively.
For all experiments, we randomly sample $20\%$ dialogues as the validation set, and others as the training set.
The model is validated every epoch, and we choose the checkpoint with the best F1 score on the validation set for evaluation.

\paragraph{Ours.}
We use OPT-13B\footnote{\url{https://huggingface.co/facebook/opt-13b}} \citep{DBLP:journals/corr/abs-2205-01068} as the LLM for generating synthetic dialogues, which is a decoder-only pre-trained language model with 13B parameters.
The learning rate and batch size are set as 1e-5 and 32. Adam optimizer \citep{DBLP:journals/corr/KingmaB14} with warmup learning rate schedule is utilized for optimization and the warmup ratio is $0.1$. 
The max training steps of LLM are 200, 400, 800 and 2K for 50, 100, 200 and 500 human dialogues respectively.
According to the performance of AutoConv on the validation set of human dialogues, we find that training LLM for 4 epochs is the most suitable. 
We randomly sample 5K documents from the training sets of QuAC and CoQA, and generate 8 synthetic dialogues for each document. 
The number of turn is set as 14 for QuAC and 30 for CoQA.
Then, we filter a quarter of the synthetic dialogues based on the diversity score of each dialogue as in \citet{DBLP:journals/corr/abs-2202-06417}, which takes into account the repetition at different $n$-gram levels.
It takes around 5 hours for training LLM and 18 hours for generating synthetic dialogues with 8 Tesla V100 32GB GPUs.

\paragraph{Evaluation.} To evaluate the quality of synthetic conversations, we evaluate the conversational question answering performance of the small task model, which is trained on both synthetic conversations and a few human conversations. The metrics are Exact Match and word-level F1 as in \citet{DBLP:conf/emnlp/ChoiHIYYCLZ18}.

\section{Baselines}
\label{app:baselines}

\paragraph{Prompting.} Prompting is a promising method for many NLP tasks.
It aims to elicit the ability of large language models learned from pre-training with text demonstrations (e.g., task instruction and few-shot examples etc).
In Table \ref{tab:main_results}, we report the results from \citet{DBLP:conf/nips/BrownMRSKDNSSAA20}.

\paragraph{Finetuning.} Train the small task model with only human annotations.

\paragraph{EDA.} Easy Data Augmentation (EDA) is a simple but effective method for text classification \citep{DBLP:conf/emnlp/WeiZ19}.
Given an input text, including both the knowledge paragraph and dialogue history in our experiments, four operations are applied to create new examples, including synonym replacement, random insertion, random swap and random deletion. 
We use their open source code\footnote{\url{https://github.com/jasonwei20/eda\_nlp}} for implementation.

\paragraph{Back-Translation.} Back-Translation is one of the most popular augmentation method for NLP tasks \cite{DBLP:conf/acl/SennrichHB16, DBLP:conf/nips/XieDHL020}. 
Specifically, we first translate the input text to a target language, then translate it back to the source language, thus we can get a paraphrased example.
To get various augmentations for each sample, we use five target languages, including Chinese, French, German, Arabic, and Korean.
Huawei Translate\footnote{\url{https://www.huaweicloud.com/product/nlpmt.html}} is used for the translation process.

\paragraph{Utterance Manipulation.} \citet{DBLP:conf/emnlp/ChenY21a} propose utterance-level manipulation to perturb the discourse relations in the conversation.
Two simple operations are used: (1) random swapping, which randomly swaps two utterances to mess up the logic chain of the conversation, and (2) random deletion, which means randomly deleting an utterance to improve the discourse diversity.
We randomly select one operation for each augmentation.

\paragraph{Dialog Inpainting.} The state-of-the-art data augmentation method for conversational question answering. 
Given a document, they iteratively insert generated utterances between the consecutive sentences in the document, then the utterances and sentences can form an informative conversation \cite{DBLP:conf/icml/DaiCZARGG22}.
We randomly sample generated dialogues from their open source data\footnote{\url{https://github.com/google-research/dialog-inpainting}}.

\end{document}